\def\year{2018}\relax
\documentclass[letterpaper]{article} 
\usepackage{aaai18}  
\usepackage{times}  
\usepackage{helvet}  
\usepackage{courier}  
\usepackage[hyphens]{url}  
\usepackage{graphicx}  
\usepackage{array}
\graphicspath{ {images/} }
\usepackage{amsmath}
\usepackage{amssymb}

\usepackage{CJK}
\usepackage{color}
\usepackage{multicol}
\usepackage{multirow}
\usepackage{array}
\usepackage{balance}
\usepackage{arydshln}
\usepackage{tikz}
\usepackage{url}

\DeclareSymbolFont{extraup}{U}{zavm}{m}{n}
\DeclareMathSymbol{\varheart}{\mathalpha}{extraup}{86}
\DeclareMathSymbol{\vardiamond}{\mathalpha}{extraup}{87}

\usepackage{xspace}

\newcommand{\eg}{\emph{e.g.,}\xspace}
\newcommand{\ie}{\emph{i.e.,}\xspace}

\DeclareMathOperator*{\argmax}{arg\,max}

\begin{document}
%

\title{Translating Pro-Drop Languages with Reconstruction Models}

\author{Longyue Wang 
\\ ADAPT Centre, Dublin City University \\ {\normalsize \em longyue.wang@adaptcentre.ie} \And 
	      Zhaopeng Tu\thanks{Zhaopeng Tu is the corresponding author.\vspace{10pt}}
	      \\ Tencent AI Lab \\ {\normalsize \em zptu@tencent.com} \And
	      Shuming Shi \\ Tencent AI Lab \\ {\normalsize \em shumingshi@tencent.com}
	      \AND
	      Tong Zhang \\ Tencent AI Lab \\ {\normalsize \em bradymzhang@tencent.com} \And
	      Yvette Graham	\\ ADAPT Centre, Dublin City University \\ {\normalsize \em yvette.graham@adaptcentre.ie} \And
	      Qun Liu	\\ ADAPT Centre, Dublin City University \\ {\normalsize \em qun.liu@adaptcentre.ie}}


\maketitle
\begin{abstract}
\begin{quote}
Pronouns are frequently omitted in pro-drop languages, such as Chinese, generally leading to significant challenges with respect to the production of complete translations. 
To date, very little attention has been paid to the dropped pronoun (DP) problem within neural machine translation (NMT).
In this work, we propose a novel reconstruction-based approach to alleviating DP translation problems for NMT models. Firstly, DPs within all source sentences are automatically annotated with parallel information extracted from the bilingual training corpus. Next, the annotated source sentence is reconstructed from hidden representations in the NMT model. 
With auxiliary training objectives, in terms of reconstruction scores, the parameters associated with the NMT model are guided to produce enhanced hidden representations that are encouraged as much as possible to embed annotated DP information. 
Experimental results on both Chinese--English and Japanese--English dialogue translation tasks show that the proposed approach significantly and consistently improves translation performance
over a strong NMT baseline, which is directly built on the training data annotated with DPs.
\end{quote}
\end{abstract}

\section{Introduction}

In pro-drop languages, such as Chinese and Japanese, pronouns can be omitted from sentences when it is possible to infer the referent from the context. 
When translating sentences from a pro-drop language to a non-pro-drop language (\eg Chinese to English), machine translation systems generally fail to translate invisible dropped pronouns (DPs).
This problem is especially severe in informal genres such as dialogues and conversation, where pronouns are more frequently omitted to make utterances more compact~\cite{yang2015recovering}. For example, our analysis of a large Chinese--English dialogue corpus showed that around 26\% of pronouns were dropped from the Chinese side of the corpus. 
This high proportion within informal genres shows the importance of addressing the challenge of translation of dropped pronouns.

\begin{CJK}{UTF8}{gbsn}
\begin{table}[t]
\renewcommand\arraystretch{1.3}
\centering
    \begin{tabular}{c|l}
        \hline
    	Input & {\bf \color{blue} (它)} 根本 没 那么 严重\\
    	Ref & {\bf \color{blue} It} is not that bad \\ \hline
    	SMT & Wasn 't that bad \\
    	NMT & {\bf \color{blue}It} 's not that bad \\ \hline
    	\hline
    	Input & 这块 面包 很 美味 ! {\bf 你} 烤 的 {\bf \color{blue} (它)} 吗 ?\\
    	Ref & The bread is very tasty ! Did you bake {\bf \color{blue}it} ? \\ \hline
    	SMT & This bread , delicious ! Did you bake ? \\
    	NMT & The bread is delicious ! Are you baked ? \\ \hline
    \end{tabular}
	\caption{Examples of translating DPs where words in brackets are dropped pronouns that are invisible in decoding. NMT model's successes on translating simple dummy pronoun (upper panel), while fails on a more complicated one (bottom panel); SMT model fails in both cases.}
	\label{tab-example}
\end{table}
\end{CJK}

Researchers have investigated methods of alleviating the DP problem for conventional Statistical Machine Translation (SMT) models showing promising results~\cite{Nagard:2010:ACL,xiang2013enlisting,wang2016naacl}. 
Modeling DP translation for the more advanced Neural Machine Translation (NMT) models, however, has received substantially less attention, resulting in low performance in this respect even for state-of-the-art approaches. 
NMT models, due to their ability to capture semantic information with distributed representations, currently only
manage to successfully translate some simple DPs, but still fail when translating anything more complex.
Table~\ref{tab-example} includes typical examples of when our strong baseline NMT system fails to accurately translate dropped pronouns.
In this paper, we narrow the gap between correct DP translation for NMT models to improve translation quality for pro-drop languages with advanced models.

More specifically, we propose a novel reconstruction-based approach to alleviate DP problems for NMT.
Firstly, we explicitly and automatically label DPs for each source sentence in the training corpus using alignment information from the parallel corpus~\cite{wang2016naacl}. Accordingly, each training instance is represented as a triple ({$\bf x$}, {$\bf y$}, {$\bf \hat{x}$}), where {$\bf x$} and {$\bf y$} are source and target sentences, and {$\bf \hat{x}$} is the labelled source sentence. Next, we apply a standard encoder-decoder NMT model to translate {$\bf x$}, and obtain two sequences of hidden states from both encoder and decoder.
This is followed by introduction of an additional {\em reconstructor}~\cite{tu2017neural} to reconstruct back to the labelled source sentence {$\bf \hat{x}$} with hidden states from either encoder or decoder, or both components. 
The central idea behind is to guide the corresponding hidden states to embed the recalled source-side DP information and subsequently to help the NMT model generate the missing pronouns with these enhanced hidden representations. 
To this end, the reconstructor produces a {\em reconstruction loss}, which measures how well the DP can be recalled and serves as an auxiliary training objective. 
Additionally, the likelihood score produced by the standard encoder-decoder measures the quality of general translation and the reconstruction score measures the quality of DP translation, and linear interpolation of these scores is employed as an overall score for a given translation.

Experiments on a large-scale Chinese--English corpus show that the proposed approach significantly improves translation performance by addressing the DP translation problem. Furthermore, when reconstruction is applied only in training, it improves parameter training by producing better hidden representations that embed the DP information.
Results show improvement over a strong NMT baseline system of +1.35 BLEU points without any increase in decoding speed. When additionally applying reconstruction during testing, we obtain a further +1.06 BLEU point improvement with only a slight decrease in decoding speed of approximately 18\%.
Experiments for Japanese--English translation task show a significant improvement of 1.29 BLEU points, demonstrating the potential universality of the proposed approach across language pairs.

\paragraph{Contributions} Our main contributions can be summarized as follows:
\begin{enumerate}
  \item We show that although NMT models advance SMT models on translating pro-drop languages, there is still large room for improvement;
  \item We introduce a reconstruction-based approach to improve dropped pronoun translation;
  \item We release a large-scale bilingual dialogue corpus, which consists of 2.2M Chinese--English sentence pairs.\footnote{Our released corpus is available at \url{https://github.com/longyuewangdcu/tvsub}.}
\end{enumerate}

\section{Background}

\subsection{Pro-Drop Language Translation}

\begin{CJK}{UTF8}{gbsn}
A pro-drop language is a language in which certain classes of pronouns are omitted to make the sentence compact yet comprehensible when the identity of the pronouns can be inferred from the context. 
Since pronouns contain rich anaphora knowledge in discourse and the sentences in dialogue are generally short, DPs not only result in missing translations of pronouns, but also harm the sentence structure and even the semantics of output. Take the second case in Table~\ref{tab-example} as an example, when the object pronoun ``它'' is dropped, the sentence is translated into ``Are you baked?'', while the correct translation should be ``Did you bake it?''. Such omissions may not be problematic for humans since they can easily recall missing pronouns from the context. They do, however, cause challenges for machine translation from a source pro-drop language to a target non-pro-drop language, since translation of such dropped pronouns generally fails.
\end{CJK}

\begin{table}[t]    
	\begin{center}
		\begin{tabular}{c|c|cc|c}
			\bf Genres & \bf Sents & \bf ZH-Pro & \bf EN-Pro & \bf DP \\
			\hline
			Dialogue & 2.15M & 1.66M & 2.26M & 26.55\% \\
			Newswire & 3.29M & 2.27M & 2.45M & 7.35\% \\ 
		\end{tabular}
	\end{center}
	\caption{\label{tab:1} Extent of DP in different genres. The {\em Dialogue} corpus consists of subtitles extracted from movie subtitle websites; The {\em Newswire} corpus is CWMT2013 news data.}
\end{table}

\begin{CJK}{UTF8}{gbsn}
As shown in Table~\ref{tab:1}, we analyzed two large Chinese--English corpora and found that around 26.55\% of English pronouns can be dropped in the dialogue domain, while only 7.35\% of pronouns were dropped in the newswire domain.
DPs in formal text genres (\eg newswire) are not as common as those in informal genres (\eg dialogue), and the most frequently dropped pronouns in Chinese newswire is the third person singular 它 (``it'') ~\cite{Baran2012AnnotatingDP}, which may not be crucial to translation performance. As the dropped pronoun phenomenon is more prevalent in informal genres, we test our method with respect to the dialogue domain.

\end{CJK}

\subsection{Encoder-Decoder Based NMT}

Neural machine translation~\cite{sutskever2014sequence,bahdanau2015neural} has greatly advanced state-of-the-art within machine translation. Encoder-decoder architecture is now widely employed, where the encoder summarizes the source sentence ${\bf x}=x_1, x_2, \dots, x_J$ into a sequence of hidden states $\{{\bf h}_1, {\bf h}_2, \dots, {\bf h}_J\}$. Based on the encoder-side hidden states, the decoder generates the target sentence ${\bf y}=y_1, y_2, \dots, y_I$ word by word with another sequence of decoder-side hidden states $\{{\bf s}_1, {\bf s}_2, \dots, {\bf s}_I\}$:
\begin{eqnarray}
P({\bf y}|{\bf x}) = \prod_{i=1}^{I} P(y_i| y_{<i}, {\bf x}) = \prod_{i=1}^{I} g(y_{i-1}, {\bf s}_i, {\bf c}_i)
\end{eqnarray}
where $g(\cdot)$ is a softmax layer. The decoder hidden state ${\bf s}_i$ at step $i$ is computed as
\begin{equation}
{\bf s}_i = f(y_{i-1}, {\bf s}_{i-1}, {\bf c}_i)
\label{eqn-hidden-state}
\end{equation}
where $f(\cdot)$ is an activation function.
${\bf c}_i$ is a weighted sum of encoder hidden states ${\bf c}_t = \sum_{j=1}^{J} \alpha_{t,j} {\bf h}_j$, where $\alpha_{t,j}$ is the alignment probability calculated by an attention model~\cite{bahdanau2015neural,Luong2015}.
The parameters of the NMT model are trained to maximize the likelihood of a set of training examples $\{[\mathbf{x}^n,\mathbf{y}^n]\}_{n=1}^N$:
\begin{eqnarray}
\mathcal{L}(\theta) = \argmax_{\theta} \sum_{n=1}^{N} \log P({\bf y}^n|{\bf x}^n; \theta)
\end{eqnarray}

Ideally, the hidden states (either encoder-side or decoder-side) should embed the missing DP information by learning the alignments between bilingual pronouns from the training corpus. In practice, however, complex DPs are still not translated correctly, as shown in Table~\ref{tab-example}. Table~\ref{tab-oracle} shows empirical results to validate this assumption. We make the following two observations: (1) the NMT model indeed outperform SMT model when translating pro-drop languages; and (2) the performance of the NMT model can be further improved by improving translation of DPs. In this work, we propose to improve DP translation by guiding hidden states to embed the missing DP information.

\begin{table}[t]
	\begin{center}
		\begin{tabular}{c|cc|c}
			\bf System & \bf Baseline & \bf Oracle & \bf $\bigtriangleup$ \\
			\hline
			SMT & 30.16 & 35.26 & +5.10 \\ 
			NMT & 31.80 & 36.73 & +4.93 \\ 
		\end{tabular}
	\end{center}
	\caption{\label{tab-oracle} Translation performance improvement (``$\bigtriangleup$'') with manually labelled DPs (``Oracle'').}
\end{table}

\section{Approach}

In the following, we discuss methods of extending NMT models with a \emph{reconstructor} to improve DP translation, which is inspired by ``reconstruction'' -- a standard concept in auto-encoder~\cite{Bourlard:1988:BC,Vincent:2010:JLMR,socher2011semi}, and successfully applied to NMT models~\cite{tu2017neural} recently.



\subsection{Architecture}

\subsubsection{Reconstructor}


The basic idea of our approach is to reconstruct the labelled source sentence from the latent representations of the NMT model and use the reconstruction score to measure how well the DPs can be recalled from latent representations. With the reconstruction score as an auxiliary training objective, we aim to encourage the latent representations to embed DP information, and thus recall the DP translation with enhanced representations.

The reconstructor reads a sequence of hidden states and the labelled source sentence, and outputs a reconstruction score. It employs an attention model~\cite{bahdanau2015neural,Luong2015} to reconstruct the labelled source sentence $\hat{\bf x}=\{\hat{x}_1, \hat{x}_2, \dots, \hat{x}_{J'}\}$ word by word, which is conditioned on the input latent representations ${\bf v}=\{{\bf v}_1, {\bf v}_2, \dots, {\bf v}_T\}$. 
The reconstruction score is computed by
\begin{eqnarray}
R(\hat{\bf x}|{\bf v}) = \prod_{j=1}^{J'} R(\hat{x}_j| \hat{x}_{<j}, {\bf v}) = \prod_{j=1}^{J'} g_r(\hat{x}_{j-1}, \hat{\bf s}_j, \hat{\bf c}_j) 
\end{eqnarray}
where $\hat{\bf s}_j$ is the hidden state in the reconstructor, and computed by
\begin{eqnarray}
\hat{\bf s}_j &=& f_r(\hat{x}_{j-1}, \hat{\bf s}_{j-1}, \hat{\bf c}_j)
\end{eqnarray}
Here $g_r(\cdot)$ and $f_r(\cdot)$ are respective softmax and activation functions for the reconstructor. The context vector $\hat{\bf c}_j$ is computed as a weighted sum of hidden states ${\bf v}$
\begin{equation}
\hat{\bf c}_j = \sum_{t=1}^{T}{\hat{\alpha}_{j,t}\cdot {\bf v}_t}
\end{equation}
where the weight $\hat{\alpha}_{j,t}$ is calculated by an additional attention model. The parameters related to the attention model, $g_r(\cdot)$, and $f_r(\cdot)$ are independent of the standard NMT model. The labeled source words $\hat{\bf x}$ share the same word embeddings with the NMT encoder.

\subsubsection{Reconstructor-Augmented NMT}

\begin{figure}[t]
\centering
\includegraphics[width=0.45\textwidth]{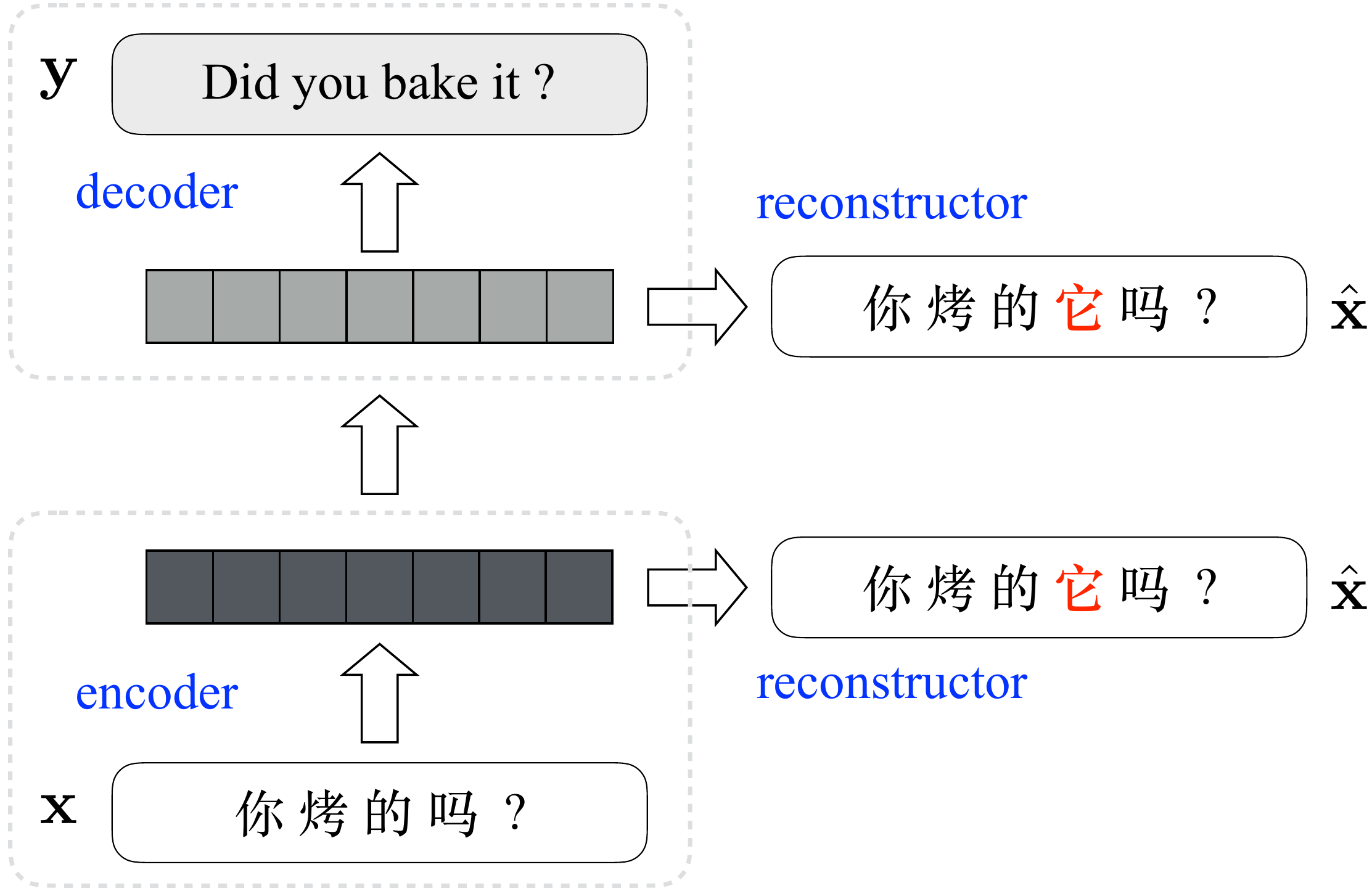}
\caption{Architecture of reconstructor-augmented NMT. The two independent reconstructors reconstruct the labelled source sentence from hidden states in the encoder and decoder, respectively.}
\label{fig-nmt-rec}
\end{figure}

We augment the standard encoder-decoder based NMT model with the introduced reconstructor, as shown in Figure~\ref{fig-nmt-rec}. The standard encoder-decoder reads the source sentence ${\bf x}$ and outputs its translation ${\bf y}$ along with the likelihood score. We introduce two independent reconstructors with their own parameters, each of which reconstructs the labelled source sentence $\hat{\bf x}$ from the encoder and decoder hidden states, respectively. 

Note that we can append only one reconstructor to either the encoder or decoder:
\begin{itemize}
  \item {\bf (encoder-reconstructor)-decoder}: When adding a reconstructor to the encoder side only, we replace the standard encoder with an enhanced {\em auto-encoder}. In the case of auto-encoding, the encoder hidden states are not only used to summarize the original source sentence but also to embed the recalled DP information from the labelled source sentence. 
  \item {\bf encoder-(decoder-reconstructor)}: This is analogous to the framework proposed by~\citeauthor{tu2017neural}~\shortcite{tu2017neural}, except that we reconstruct back to the labelled source sentence rather than the original one. It encourages the decoder hidden states to embed complete information from the source side, including the recalled DPs.
\end{itemize}
As seen, reconstructors applied on different sides of the corpus may capture different patterns of DP information, and using them together can encourage both the encoder and decoder to learn recalled DP information.
Our approach is very much inspired by recent successes within question-answering, where a single information source is fed to multiple memory layers so that new evidence is captured in each layer and combined into subsequent layers~\cite{Sukhbaatar:2015:NIPS,Miller:2016:EMNLP}.

\begin{table*}[t]
\centering
\renewcommand\arraystretch{1.3}
\begin{tabular}{c|c cc c cc c cc c cc}
\multirow{2}{*}{\bf Data}	&	\multirow{2}{*}{$|S|$}	&	\multicolumn{2}{c}{$|W|$} & & \multicolumn{2}{c}{$|P|$} & 	&	\multicolumn{2}{c}{$|V|$} & & 	\multicolumn{2}{c}{$|L|$} \\
\cline{3-4}\cline{6-7}\cline{9-10}\cline{12-13}
				&	&	Zh	&	En	&	&	Zh	&	En	&	&	Zh	&	En	&	&	Zh	&	En\\
\hline

Train	&	2.15M   & 12.1M  &  16.6M &   &   1.66M   &   2.26M   &   & 151K  & 90.8K  &   &   5.63    &   7.71\\
Tune	&	1.09K   & 6.67K  & 9.25K  &   &   0.76K   &   1.03K   &   &  1.74K & 1.35K  &   &   6.14    &   8.52\\
Test	&	1.15K   & 6.71K  & 9.49K  &   &   0.77K   &   0.96K   &   &  1.79K & 1.39K  &   &   5.82    &   8.23\\
\end{tabular}
\caption{Number of sentences ($|S|$), words ($|W|$), pronouns ($|P|$), vocabulary ($|V|$), and averaged sentence length ($|L|$) comprising the training, tuning and test corpora. K stands for thousands and M for millions.}
\label{table-statistics}
\end{table*}

\subsection{Training and Testing}
\subsubsection{Training} We train both the encoder-decoder and the introduced reconstructors together in a single end-to-end process. The two-reconstructor model (Figure~\ref{fig-nmt-rec}) are described below (the other two individual models correspond to each part). The training objective can be revised as
\begin{eqnarray}
J(\theta, \gamma, \psi) = \argmax_{\theta, \gamma, \psi} \sum_{n=1}^{N} \bigg\{ \underbrace{\log P({\bf y}^n|{\bf x}^n; \theta)}_\text{\normalsize \em likelihood} \nonumber \\ 
 + \underbrace{\log R_{enc}({\bf \hat{x}}^n | {\bf h}^n; \theta, \gamma)}_\text{\normalsize \em enc-rec} + \underbrace{\log R_{dec}({\bf \hat{x}}^n | {\bf s}^n; \theta, \psi)}_\text{\normalsize \em dec-rec} \bigg\}
\end{eqnarray}
where $\theta$ is the parameter matrix in encoder-decoder, and $\gamma$ and $\psi$ are model parameters related to the {\em encoder-side reconstructor} (``enc-dec'') and {\em decoder-side reconstructor} (``dec-rec'') respectivley; ${\bf h}$ and ${\bf s}$ are encoder and decoder hidden states. The auxiliary reconstruction objectives (\eg $R_{enc}(\cdot)$ and $R_{dec}(\cdot)$) guide the related part of the parameter matrix $\theta$ to learn better latent representations, which are used to reconstruct 
the labelled source sentence.

\subsubsection{Testing}

In testing, reconstruction can serve as a reranking technique to select a better translation from the $k$-best candidates generated by the decoder. Each translation candidate is assigned a likelihood score from the standard encoder-decoder, as well as reconstruction score(s) from the newly added reconstructor(s). Since the target sentence is invisible in testing, we employ a monolingual labelling model built on the training corpus to label DPs in the input sentence~\cite{wang2016naacl}.

When using reconstruction in testing, it requires external resources (\ie monolingual DP label tool) and more computations (\ie calculation of reconstruction scores). To reduce the dependency and cost, we can also employ a standard encoder-decoder model with better trained parameters so that the parameters can produce enhanced latent representations that embed DP information. Such information is invisible in the original input sentence but can be learned from the training data with similar context.

\section{Experiments}

\subsection{Data}
Experiments evaluate the method for translation of Chinese--English subtitles. More than two million sentence pairs 
were extracted from the subtitles of television episodes.\footnote{The data were crawled from the subtitle website \url{http://www.zimuzu.tv}.} 
We pre-processed the extracted data using our in-house scripts~\cite{wang2016lrec2}, including sentence boundary detection and bilingual sentence alignment etc. Finally, we obtained a high-quality corpus which keeps the discourse information.
Table~\ref{table-statistics} lists the statistics of the corpus. 
Within the subtitle corpus, sentences are generally short and the Chinese side, as expected, contains many examples of dropped pronouns. We randomly select two complete television episodes as the tuning set, and another two episodes as the test set. 
We used case-insensitive 4-gram NIST BLEU metrics \cite{Papineni:2002} for evaluation, and {\em sign-test} \cite{Collins05} to test for statistical significance.

\subsection{DP Annotation}

We follow~\citeauthor{wang2016naacl}~\shortcite{wang2016naacl} to automatically label DPs for training and test data. In the {\em training phase}, where the target sentence is available, we label DPs for the source sentence using alignment information. These labeled source sentences can be used to build a monolingual DP generator using NN, which is used to label test sentences since the target sentence is not available during the {\em testing phase}. The F1 scores of the two approaches on our data are 92.99\% and 65.21\%, respectively. After automatic labelling, the number of pronouns on the Chinese side in training, tuning and test data are 2.09M, 0.98K, 0.96K respectively, which is roughly consistent with pronoun frequency on the English side.

The usage of the labeled source sentences is two-fold:
\begin{itemize}
    \item [1.] {\em Baseline (+DPs)}: a stronger baseline system trained on the new parallel corpus (labelled source sentence, target sentence), which is evaluated on the new test sentences labelled by the monolingual DP generator.
    \item [2.] {\em Our models}: the proposed models reconstruct hidden states back to the labelled source sentences. 
\end{itemize}
For the source sentences that have no DPs, we use the original ones as labelled source sentences, otherwise we use the DP-labeled sentences.


\begin{table*}[t]
\centering
\renewcommand\arraystretch{1.3}
\begin{tabular}{l|c|c|c|l|c}
	\multirow{2}{*}{\bf Model}               & \multirow{2}{*}{\bf \#Params}  &  \multicolumn{2}{c|}{\bf Speed}  &   \multicolumn{2}{c}{\bf BLEU}\\
	\cline{3-6}
	    & &   Training    &   Decoding    &  Test &  $\bigtriangleup$ \\
	\hline
	Baseline                &  86.7M &   1.60K   &  2.61    & 31.80 &   -- / -- \\   
	Baseline (+DPs)         &  86.7M  &   1.59K    &   2.63   & 32.67$^{\dag}$ & +0.87 / -- \\
	\hline
	~~+ enc-rec             & +39.7M  &   0.71K   &   2.63   & 33.67$^{\dag\ddag}$ & +1.87 / +1.00 \\ 
	~~+ dec-rec             & +34.1M  &   0.84K   &   2.18   & 33.48$^{\dag\ddag}$ & +1.68 / +0.81 \\
	~~+ enc-rec + dec-rec   & +73.8M  &   0.57K   &   2.16   & {\bf 35.08}$^{\dag\ddag}$ & {\bf +3.28} / {\bf +2.41} \\ 
	\hline \hline
	Multi-Source~\cite{DBLP:conf/naacl/ZophK16}            & +20.7M  &     1.17K      &  1.27 & 32.81$^{\dag}$ & +1.01 / +0.14 \\
    {Multi-Layer~\cite{wu2016google}}    &  +75.1M  &   0.61K    &   2.42   & 33.36$^{\dag}$ & +1.56 / +0.69 \\ 
    Baseline (+DPs) + Enlarged Hidden Layer      &  +86.6M  &   0.68K    &   2.51   & 32.00$^{\dag}$ & +0.20 / -0.67 \\ 
\end{tabular}
\caption{\label{tab-results} Evaluation of translation performance for Chinese--English. ``Baseline'' is trained and evaluated on the original data, while ``Baseline (+DPs)'' is trained on the data labelled with DPs. ``enc-rec'' indicates encoder-side reconstructor and ``dec-rec'' denotes decoder-side reconstructor.
Training speed is measured in words/second and decoding speed is measured in sentences/second with beam size being 10. The two numbers in the ``$\bigtriangleup$'' column denote performance improvements over ``Baseline'' and ``Baseline (+DPs)'', respectively.
``$\dag$'' and ``$\ddag$'' indicate statistically significant difference ($p < 0.01$) from ``Baseline'' and ``Baseline (+DPs)'', respectively. All listed models except ``Baseline'' exploit the labelled source sentences.
}
\end{table*}

\subsection{Model}
The baseline is our re-implemented attention-based NMT system, which incorporates dropout \cite{hinton2012improving} on the output layer and improves the attention model by feeding the most recently generated word. For training the baseline models, we limited the source and target vocabularies to the most frequent 30K words in Chinese and English, covering approximately 97.2\% and 99.3\% of the words in the two languages, respectively. Each model was trained on sentences of length up to a maximum of 20 words with early stopping. Mini-batches were shuffled during processing with a mini-batch size of 80. The word-embedding dimension was 620 and the hidden layer size was 1,000. We trained for 20 epochs using Adadelta~\cite{zeiler2012adadelta}, and selected the model that yielded best performances on the tuning set.

The proposed model was implemented on top of the baseline model with the same settings where applicable. The hidden layer size in the reconstructor was 1,000. Following~\citeauthor{tu2017neural}~\shortcite{tu2017neural}, we initialized the parameters of our models (\ie encoder and decoder, except those related to reconstructors) with the baseline model. We further trained all the parameters of our model for another 15 epochs.


\subsection{Results and Discussion}

Table~\ref{tab-results} shows translation performances for Chinese--English.
Clearly the proposed models significantly improve the translation quality in all cases, although there are still considerable differences among different variants.


\paragraph{Baselines} The two baseline NMT models, one being trained and evaluated on the original bilingual data without any explicitly labelled DPs (\ie ``Baseline''), while the other was trained and evaluated on the labelled data (\ie ``Baseline (+DPs)''). As can be seen from the BLEU scores, the latter significantly outperforms the former, indicating that explicitly recalling translation of DPs helps produce better translations. 
Benefiting from the explicitly labelled DPs, the stronger baseline system is able to improve performance over the standard baseline system built on the original data where the pronouns are missing.

\paragraph{Parameters} In terms of additional parameters introduced by the reconstruction models, both reconstructors introduce a large number of parameters. Beginning with the baseline model's 86.7M parameters, the encoder-side reconstructor adds 39.7M new parameters, while the decoder-side reconstructor adds a further 34.1M new parameters. Furthermore, adding reconstructors to both sides leads to additional 73.8M parameters. 
More parameters may capture more information, at the cost of posing difficulties to training.

\paragraph{Speed} Although gains are made in terms of translation quality by introducing reconstruction, we need to consider the potential trade-off with respect to a possible increase in training and decoding times, due to the large number of newly introduced parameters resulting from the incorporation of reconstructors into the NMT model. When running on a single GPU device Tesla K80, the training speed of the baseline model is 1.60K target words per second, and this reduces to 0.57K words per second when reconstructors are added to both sides. 
In terms of decoding time trade-off, our most complex model only decreases decoding speed by 18\%. We attribute this to the fact that no beam search is required for calculating reconstruction scores, which avoids the very costly data swap between GPU and CPU memories.

\paragraph{Translation Quality}
Clearly the proposed approach significantly improves the translation quality in all cases, although there are still considerable differences among the proposed variants. Introducing encoder-side and decoder-side reconstructors individually improves translation performance over ``Baseline (+DPs)'' by +1.0 and +0.8 BLEU points respectively. Combining them together achieves the best performance overall, which is +2.4 BLEU points better than the strong baseline model. This confirms our assumption that reconstructors applied to the source and target sides indeed capture different patterns for translating DPs.

\paragraph{Comparison to Other Work}

For the purpose of comparison, we reimplemented the multi-source model of~\citeauthor{DBLP:conf/naacl/ZophK16}~\shortcite{DBLP:conf/naacl/ZophK16}, which introduces an alternate encoder (shared parameters) and attention model (independent parameters) that take labelled sentences as an additional input source. This multi-source model significantly outperforms our ``Baseline'' model without labelled DP information, but only marginally outperform the ``Baseline (+DPs)'' that uses labelled DPs. One possible reason is that the two sources (\ie original input and labelled input sentences) are too similar to one another, making it difficult to distinguish them from labelled DPs. 

Some may argue that the BLEU improvements are mainly due to the model parameter increase (\eg +73.8M) or deeper layers (\eg two reconstruction layers). To answer this concern, we compared the following two models:
\begin{itemize}
    \item Multi-Layer~\cite{wu2016google}: a system with three-layer encoder and three-layer decoder. The additional layers introduce 75.1M parameters, which is in the same scale with the proposed model (\ie 73.8M).
    \item Baseline (+DPs) + Enlarged Hidden Layer: a system with the same setting as ``Baseline (+DPs)'' except that layer size is 2100 instead of 1000. This variant introduces 86.6M parameters, which is even more than the most complicated variant of proposed models.
\end{itemize}
We found that the multi-layer model significantly outperforms its single-layer counterpart ``Baseline (+DPs)'', while significantly underperforms our best model (\ie 33.46 vs. 35.08). The ``Baseline (+DPs)'' system with enlarged hidden layer, however, does not achieve any improvement. This indicates that explicitly modeling DP translation is the key factor to the performance improvement.


\paragraph{Japanese--English Translation Task}

\begin{table}[h]
\centering
\renewcommand\arraystretch{1.1}
\begin{tabular}{l|c|c}
	\bf Model & \bf Test & \bf $\bigtriangleup$ \\
	\hline
	Baseline (+DPs)     &   20.55   &   --\\
	\hline
	~~+ enc-rec + dec-rec & 21.84   &   + 1.29\\
\end{tabular}
\caption{\label{tab-results-jaen} Evaluation of translation performance for Japanese--English.}
\end{table}

To validate the robustness of our approach on other pro-drop languages, we conducted experiments on Opensubtitle2016\footnote{\url{http://opus.nlpl.eu/OpenSubtitles2016.php}} data for the Japanese--English translation. 
We used the same settings as used in Chinese--English experiments, except that the vocabulary size is 20,001. 
As shown in Table \ref{tab-results-jaen}, our model also significantly improves translation performance on the Japanese--English task, demonstrating the efficiency and potential universality of the proposed approach.

\subsection{Analysis}

We conducted extensive analyses for Chinese--English translation to better understand our model in terms of contribution of reconstruction from training and testing, effect of reconstructed input, effect of DP labelling accuracy, and building the ability to handling long sentences.

\subsubsection{Contribution Analysis}

\begin{table}[h]
\centering
\renewcommand\arraystretch{1.1}
\begin{tabular}{l|c|c}
	\bf Model & \bf Test & \bf $\bigtriangleup$ \\
	\hline
	Baseline            &   31.80   &   -- / --\\
	Baseline (+DPs)     &   32.67   &   +0.87 / --\\
	\hline
	~~+ enc-rec         &   33.67   &   +1.87 / +1.00\\
	~~+ dec-rec         &   33.15   &   +1.35 / +0.48\\
	~~+ enc-rec + dec-rec & {\bf 34.02}   &   {\bf +2.22} / {\bf +1.35}\\
\end{tabular}
\caption{\label{tab-results-no-rec} Translation results when {\em reconstruction is used in training only while not used in testing}.}
\end{table}

As mentioned previously, the effect of reconstruction is two-fold: (1) it improves the training of baseline parameters, which leads to better hidden representations that embed labelled DP information learned from the training data; and (2) it serves as a reranking metric in testing to measure the quality of DP translation.\footnote{As in testing encoder-side reconstructor reconstructs back to the same labelled source sentence with the same encoder hidden states, all translation candidates would share the same encoder-side reconstruction score. Therefore, in such cases, reconstruction cannot be used as a reranking metric.} Table~\ref{tab-results-no-rec} lists translation results when the reconstruction model is used in training only. Results show all variants to outperform the baseline models and applying reconstructors to both sides achieves the best performance overall. This is encouraging, since no extra resources nor computation are introduced to online decoding, making the approach highly practical, for example for translation in industry applications.

\subsubsection{Effect of Reconstruction}

\begin{table}[h]
\centering
\renewcommand\arraystretch{1.1}
\begin{tabular}{l|c|c}
	\bf Model & \bf Test & \bf $\bigtriangleup$ \\
	\hline
	Baseline            &   31.80   &   -- / --\\
	Baseline (+DPs)     &   32.67   &   +0.87 / --\\
	\hline
	~~+ enc-rec         &   33.21   &   +1.41 / +0.54\\
	~~+ dec-rec         &   33.08   &   +1.28 / +0.41\\
	~~+ enc-rec + dec-rec & {\bf 33.25}   &   {\bf +1.45} / {\bf +0.58}\\
\end{tabular}
\caption{\label{tab-results-no-dp} Translation results when hidden states are {\em reconstructed into the original source sentence} instead of the source sentence labelled with DPs.}
\end{table}

Some researchers may argue that the proposed method acts much like dual learning~\cite{He:2016:NIPS} and reconstruction~\cite{tu2017neural} especially when sentences have no DPs, which can benefit to the overall translation, not just only with respect to DPs.
To investigate to what degree the improvements are indeed made by explicitly modeling DP translation, we examine the performance of variants which reconstruct hidden states back to the original input sentence instead of the source sentence labelled with DPs, as shown in Table~\ref{tab-results-no-dp}. Note that the variant ``+ dec-rec'' in this setting is exactly the model proposed by~\citeauthor{tu2017neural}~\shortcite{tu2017neural}. As seen, although the variants significantly outperforms ``Baseline'' model without using any DP information, the absolute improvements are still worse than our proposed model that explicitly exploits DP information (\ie 1.45 vs. 3.28). This validates our hypothesis that explicitly modeling DP translation contributes most to the improvement.

\subsubsection{Effect of DP Labelling Accuracy}

\begin{table}[h]
\centering
\renewcommand\arraystretch{1.1}
\begin{tabular}{l|c|c|c}
	\bf Model & \bf Automatic   &   \bf Manual   & \bf $\bigtriangleup$ \\
	\hline
	Baseline (+DPs)     &   32.67   &   36.73   &   +4.06\\
	\hline
	~~+ enc-rec         &   33.67   &   37.58   &   +3.91\\
	~~+ dec-rec         &   33.48   &   37.23   &   +3.75\\
	~~+ enc-rec + dec-rec & 35.08   &   38.38   &   +3.30\\
\end{tabular}
\caption{\label{tab-results-DP-accuracy} Translation performance gap (``$\bigtriangleup$'') between manually (``Manual'') and automatically (``Automatic'') labelling DPs for input sentences in testing.}
\end{table}

For each sentence in testing, the DPs are labelled automatically by a DP generator model, the accuracy of which is 65.21\% measured in F1 score. The labelling errors may propagate to the NMT models, and have the potential to negatively affect translation performance. We investigate this using manual labelling and automatic labelling, as shown in Table~\ref{tab-results-DP-accuracy}. The analysis firstly shows that there still exists a significant gap in performance, and this could be improved by improving the accuracy of DP generator. Secondly, our models show a relatively smaller distance in performance from the oracle performance (``Manual''), indicating that the proposed approach is more robust to labelling errors.

\subsubsection{Length Analysis}

\begin{figure}[h]
\centering
\graphicspath{ {figures/} }
\includegraphics[width=0.45\textwidth]{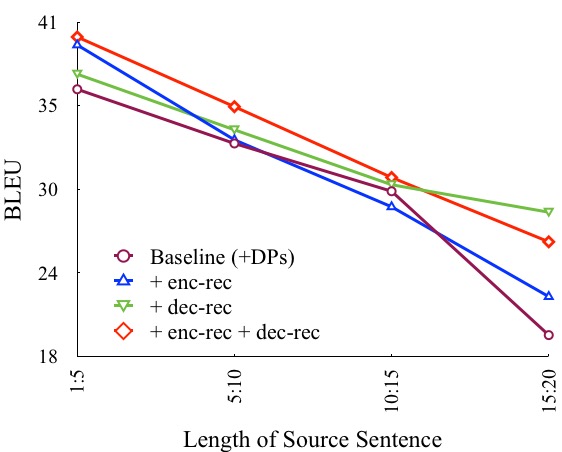}
\caption{Performance of the generated translations with respect to the lengths of the source sentences.}
\label{fig:3}
\end{figure}

Following~\cite{bahdanau2015neural,Tu:2016:ACL,Tu:2017:TACL}, 
we group sentences of similar lengths together and compute the BLEU score for each group, as shown in Figure~\ref{fig:3}. 
The proposed models outperform the baseline for most span lengths, although there are still some notable differences. 
The improvement achieved by the source-side reconstructor is mainly for translation of short sentences (\eg $<5$), while that of the target-side reconstructor is mainly for translation of long sentences (\eg $>15$). The reason is that (1) reconstruction can make encoder-side hidden states contain complete source information including DP information and subsequently good performance on short sentences, while at the same time, they cannot guarantee that all the information will be transferred to the decoder side (\ie relatively bad performance on long sentences); (2) similar to findings of ~\cite{tu2017neural}, the decoder-side reconstructor can make translation more adequate, which significantly alleviates inadequate translation problems for longer sentences. Combining them together can take advantage of both models, and thus the improvements are more substantial for all span lengths.

\paragraph{Error Analysis}

\begin{table}[h]
\renewcommand\arraystretch{1.1}
\centering
\begin{tabular}{l|c|ccc|c}
 {\bf Model}    &   \bf Error & \bf Sub. & \bf Obj. & \bf Dum. & \bf All\\ 
 \hline
 \textsc{Base}  & Total  & 112 & 41 & 45 & 198 \\ 
 \hline
 \multirow{2}{*}{\textsc{+ enc}}  & Fixed  & 51 & 22 & 28  & 101 \\
 &  New    & 25 & 8 & 4 & 37 \\ \hline
 \multirow{2}{*}{\textsc{+ dec}}  & Fixed  & 57 & 21 & 17  & 95 \\
 &  New    & 19 & 10 & 6 & 36 \\ \hline
  \multirow{2}{*}{\textsc{+ enc + dec}}  & Fixed  & 50 & 34 & 33  & 117 \\
 &  New    & 11 & 14 & 7 & 32 \\
\end{tabular}
\caption{\label{tab:8} Translation error statistics on different types of pronouns: subject (``Sub.''), object (``Obj.'') and dummy (``Dum.'') pronouns. ``\textsc{Base}'' denotes ``Baseline (+DPs)'', ``\textsc{+ enc}'' denotes ``+ enc-rec'', ``\textsc{+ dec}'' denotes ``+ dec-rec'' and ``\textsc{+ enc + dec}'' denotes ``+ enc-rec + dec-rec''.}
\end{table}

We investigate to what extent DP-related errors are fixed by the proposed models. We randomly select 500 sentences from the test set and count errors produced by the strong baseline model (``Total''), what proportion of these are fixed (``Fixed'') or newly introduced (``New'') by our approach, as shown in Table~\ref{tab:8}. All the proposed models can fix different kinds of DP problems, and the ``\textsc{+ ENC + DEC}'' variant achieves the best performance, which is consistent with the translation results reported above.
The ``\textsc{+ ENC + DEC}'' model fixed 59.1\% of the DP-related errors, while only introducing 16.2\% of new errors. This confirms that our improvement in terms of automatic metric scores indeed comes from alleviating DP translation errors.

Among all types of pronouns, translation errors on object and dummy pronouns,\footnote{A dummy pronoun (\ie ``it'') is a pronoun used for syntax without explicit meaning. It is used in Germanic languages such as English but not in Pro-drop languages such as Chinese.} which can be usually inferred with intra-sentence context, are easy to be alleviated. In contrast, errors related to the subject of a given sentence are more difficult, since labelling dropped pronouns in such cases generally requires cross-sentence context. Table~\ref{fig:8} shows three typical examples of successfully fixed, failed to fix, and newly introduced subjective-case pronouns.

\begin{CJK}{UTF8}{gbsn}
\begin{table}[h]
\renewcommand\arraystretch{1.3}
\centering
\begin{tabular}{c|l}
\hline
\multicolumn{2}{c}{\bf Fixed Error} \\
\hline
Input & 等 我 搬进 来 {\bf (我)} 可以 买 一台 泡泡机 吗 ?\\ 
Ref. & When I move in, can I get a bubble machine?\\ 
NMT & When I move in {\em \color{blue} to} buy a bubble machine.\\ 
Our & When I move in, can {\bf \color{red} I} buy a bubble machine?\\
\hline
\multicolumn{2}{c}{\bf Non-Fixed Error} \\
\hline
Input & {\bf (他)} 是 个 训练营 ?\\ 
Ref. &  It is a camp?\\ 
NMT & {\em \color{blue} He} was a camp?\\ 
Our & {\em \color{blue} He}'s a camp?\\
\hline
\multicolumn{2}{c}{\bf Newly Introduced Error} \\
\hline
Input & {\bf (我)} 要 把 这 戒指 还给 你\\ 
Ref. &  I need to give this ring back to you.\\ 
NMT & {\bf \color{red}I}'m gonna give you the ring back.\\ 
Our & {\em \color{blue}To} give it back to you.\\
\hline
\end{tabular}
\caption{\label{fig:8} Example translations where subjective-case pronouns in brackets are dropped in original input but labeled by DP generator. We italicize some {\em \color{blue} mis-translated} errors and highlight the {\bf \color{red} correct} ones in bold.}
\end{table}

\section{Related Work}


\paragraph{DP Translation for SMT}
Previous research has investigated DP translation for SMT. For example, \citeauthor{chung2010effects}~\shortcite{chung2010effects} examined the effects of empty category (including DPs) on MT with various methods. This work showed improvements in terms of translation performance despite the automatic prediction of empty category not being highly accurate. \citeauthor{Taira:2012:SSSST}~\shortcite{Taira:2012:SSSST} analyzed the Japanese-to-English translation by inserting DPs into input sentences using simple rule-based methods, achieving marginal improvements. More recently, \citeauthor{wang2016naacl}~\shortcite{wang2016naacl} proposed labelling DPs using parallel information of training data, and obtained promising results in SMT.~\citeauthor{Wang2017journal}~\shortcite{Wang2017journal} also extend the SMT-based DP translation method on Japanese--English translation task. Inspired by these previous successes, this paper is an early attempt to learn to tackle DP translation for NMT models.

\paragraph{Representation Learning with Reconstruction}

Reconstruction is a standard concept in auto-encoder, that guides towards learning representations that captures the underlying explanatory factors for the observed input~\cite{Bourlard:1988:BC,Vincent:2010:JLMR}. 
An auto-encoder model consists of an encoding function to compute a representation from an input, and a decoding function to reconstruct the input from the representation.
The parameters involved in the two functions are trained to maximize the {\em reconstruction score}, which measures the similarity between the original input and reconstructed input.
Inspired by the concept of {\em reconstruction},~\citeauthor{tu2017neural}~\shortcite{tu2017neural} proposed guiding decoder hidden states to embed complete source information by reconstructing the hidden states back to the original source sentence. Our approach differs at: (1) we introduced not only decoder-side reconstructor but also encoder-side reconstructor to learn enhanced hidden states of both encoder and decoder; and (2) we guide the hidden states to embed complete source information as well as the labelled DP information.

\paragraph{Multiple Sources for NMT}
Recently, it was shown that NMT can be improved by feeding auxiliary information sources beyond the original input sentence. The additional sources can be in various forms, such as parallel sentences in other languages~\cite{dong2015multi,DBLP:conf/naacl/ZophK16}, cross-sentence contexts~\cite{wang2017exploiting,jean2017does,Tu:2018:TACL}, generation recommendations from other translation models~\cite{He:2016:AAAI,Wang:2017:AAAI,Gu:2017:arXiv,WangXing:2017:EMNLP}, syntax information~\cite{Li:2017:ACL,Zhou:2017:ACL}. Along the same direction, we provide complementary information in terms of source sentences labelled with DPs.

\section{Conclusion and Future Work}

This paper is an early attempt to model DP translation for NMT systems. 
Hidden states are guided in both the encoder and decoder to embed the DP information by reconstructing them back to the source sentence labelled with DPs. The effect of reconstruction model is two-fold: (1) it improves parameter training for producing better latent representations; and (2) it measures the quality of DP translation, which is combined with likelihood to better measure the overall quality of translations. 
Quantity and quality analyses show that the proposed approach significantly improves translation performance across language pairs, and can be further improved by developing better DP labelling models.

In future work we plan to validate the effectiveness of our approach on other text genres with different prevalence of DPs.
For example, in formal text genres (\eg newswire), DPs are not as common as in the informal text genres, and the most frequently dropped pronouns in Chinese newswire is the third person singular ``它'' (``{\em it}'')~\cite{Baran2012AnnotatingDP}, which may not be crucial to translation performance.
\end{CJK}

\section{Acknowledgments}
The ADAPT Centre for Digital Content Technology is funded under the SFI Research Centres Programme (Grant 13/RC/2106) and is co-funded under the European Regional Development Fund. Work was done when Longyue Wang was interning at Tencent AI Lab.

\balance

\bibliographystyle{aaai}

\end{document}